\documentclass[paper]{ieice}
\usepackage[pdftex]{graphicx,xcolor}
\usepackage[fleqn]{amsmath}
\usepackage{newtxtext}
\usepackage[varg]{newtxmath}
\usepackage{multirow}
\usepackage{soul}
\field{A}
\title{Capsule network with shortcut routing}
\authorlist{%
 \authorentry[dtvu1707@gmail.com]{Dang Thanh Vu}{n}{labelA}\MembershipNumber{}
 \authorentry{Vo Hoang Trong}{n}{labelA}\MembershipNumber{}
 \authorentry{Yu Gwang Hyun}{n}{labelA}\MembershipNumber{}
 \authorentry{Kim Jin Young}{m}{labelA}\MembershipNumber{1385259}
}
\affiliate[affiliate label]{Department of ICT Convergence System Engineering, Chonnam National University, Republic of Korea
}

\received{2015}{1}{1}
\revised{2015}{1}{1}

\begin{document}
\maketitle
\newcommand{\norm}[1]{\left\lVert#1\right\rVert}
\begin{summary}
Capsules are fundamental informative units that are introduced into capsule networks to manipulate the hierarchical presentation of patterns. The part–whole relationship of an entity is learned through capsule layers, using a routing-by-agreement mechanism that is approximated by a voting procedure. Nevertheless, existing routing methods are computationally inefficient. We address this issue by proposing a novel routing mechanism, namely "shortcut routing", that directly learns to activate global capsules from local capsules. In our method, the number of operations in the routing procedure is reduced by omitting the capsules in intermediate layers, resulting in lighter routing. To further address the computational problem, we investigate an attention-based approach, and propose fuzzy coefficients, which have been found to be efficient than mixture coefficients from EM  routing. Our method achieves on-par classification results on the Mnist (99.52\%), smallnorb (93.91\%) , and affNist (89.02\%) datasets. Compared to EM routing, our fuzzy-based and attention-based routing methods attain reductions of $1.42$ and $2.5$ {in terms of the number of calculations}.
\end{summary}

\section{Introduction}
\subsection{Capsule network}

A  capsule is a group of neurons \cite{1} that encode the viewpoint variation of an entity. There are two quantities present in a capsule: the instantiation (pose) vector and the activation probability. The presence of an entity is disclosed by its activation probability, while its instantiation vector encodes the perturbations of that entity on a manifold. A CapsNet is made of capsule layers that are stacked on top of each other \cite{2,3}. In CapsNets, the information flows from the lower layer to the higher layer via a voting mechanism  \cite{1}.

Typical CNNs cannot perform tasks that require hierarchical parsing of a visual scene \cite{1}. Instead, a capsule is designed to encapsulate the part–whole relationship into a group of neurons, particularly the instantiation vector. Intuitively, to formulate equivariance property, one capsule encodes three information: the position, shape, and existence of a visual entity \cite{4} .

\subsection{Invariance and Equivariance}
Convolution is locally invariant to variant factors (e.g., translation, scaling, and rotation). The weight-sharing scheme and pooling mechanism also help CNNs to achieve local translational invariance. Poolings and convolutions, work as feature detectors, locally retaining the most activated neuron in the receptive fields, resulting in compact feature maps that help CNNs to recognize an object. However, those operators usually discard variant factors on the appearance manifold.

Having introduced a parse tree-like structure of objects, \cite{6} stated that a visual entity is constructed of its parts. The structural relationship of these entities is viewpoint-invariant under affine transformations, according to a reference fragment \cite{7}. This idea motivates the use of matrix multiplication to investigate the part–whole relationship and test of agreement to examine if an entity is present \cite{1,2}. In terms of topology, instantiation parameters represent the abstract variation, where each instantiation corresponds to each coordinate of the chart, mapping from the appearance manifold into the variance manifold \cite{5,8}. The capsules attain the equivariance property when their instantiations change  with respect to the viewpoint defined by the chart map.

\subsection{High computation}
However, computational complexity limits the feasibility of CapNets. This deficiency emerges both from the capsule units, and routing algorithm. As a capsule unit is a group of $n$ neurons, storing one capsule layer thus requires $n$ times more memory than storing a standard hidden layer. Moreover, the larger the capsules, the more learnable weights that are needed. Passing information through capsule layers involves matrix multiplication, which requires a vast number of operations. Furthermore, when the number of routing iterations increases, the required memory and computation time significantly increase. In general, this computational problem can be tackled in two different ways. One approach is to reduce the size of the voting tensors, while the other is to simplify the routing procedure.

\subsection{Contribution}
Our main contributions in this article can be summarized as follows:

\begin{enumerate}
	\item {To resolve the computational expense of capsule networks, we introduce routing paths that operate as shortcuts} \cite{10}, {rather than as an unfolding sequence of layers. Shortcut flows iteratively update the capsules at the last layer by a routing algorithm.} 
	
	\item We address the computational problem in the routing algorithm by introducing two simplified ways of calculating the routing coefficients: the attention-based approach based on cosine similarity, and the fuzzy-based approach based on fuzzy coefficients.
	
	\item We conduct experiments on a classification task. To evaluate the performance of our framework on datasets that include varying viewpoints, in addition to the Mnist dataset, we also experiment on the affNist and smallNorb datasets.
\end{enumerate}

The rest of this paper is organized as follows: Section 2 presents related works in the literature on capsule networks. Section 3 describes the underlying idea of shortcut routing, and formulates the mathematical equations. Section 4 presents a baseline CapsNet architecture and the experimental results. Finally, Section 5 concludes this study.

\section{Related works}
Many studies have introduced specific mechanisms to standard CNNs, or regularized the objective function to encourage CNNs to observe the invariance inside a dataset. Kulkarni et al. proposed a deep convolutional inverse graphics network that learns graphics codes corresponding to distinct transformations \cite{12}. Jaderberg et al. introduced a differentiable spatial transformer module that can produce an appropriate degradation for each input example \cite{13}. Worral et al. replaced ordinary convolutional layers with steerable circular harmonic layers to exhibit patch-wise translations and rotations \cite{14}. A widely acceptable way to increase the invariance capacity of a model is to train it with adversarial samples \cite{15,16}, which introduces a locally invariant prior, and encourages neural networks exploit in the neighborhood of a data manifold \cite{16}. The studies of Lenssen \cite{17} and Cohen \cite{18} presented equivariance under group theory, notes that capsule networks do not ensure equivariance, because there is no guarantee that the primary capsule layer will encode the spatial domain into the receptive field of the poses.

Capsule networks have been studied extensively since routing algorithms were presented \cite{2,3}. Sabour et al. developed the dynamic routing algorithm to be compatible with vector capsules whose orientation comprises the pose information, while the length of the pose vector is the activation probability. Hinton et al. proposed EM routing based on the expectation-maximization algorithm. They also applied patch-wise convolution to capsule layers. These two studies constitute the fundamental architecture of capsule networks.

Several versions of CapsNet have modified the architecture and routing algorithm \cite{30,31,32}. Zhang introduced two routing algorithms based on weighted kernel density estimation, one depends on the mean-shift algorithm, while the is a simplified version of EM routing \cite{19}. Duarte et al. proposed VideoCapsuleNet, the main contribution being capsule pooling, which takes the average of all capsules in a receptive field before routing to higher levels \cite{20}. To make CapsNets deeper, Rajasegaran et al. suggested using 3D convolutions instead of matrix multiplication, to increase the number of hidden layers; the third dimension pertains to the pose vector of capsules \cite{21}. Although the authors announced that this approach surpasses the results of standard CapsNets, their framework does not align with  the intuition of capsule units. Zhang \cite{22} formulated learning multiple capsule subspaces by orthogonal projection. This projection can be embedded into the primary capsule layer to explore correlations between feature maps and capsule subspaces. Lalonde et al. revealed deconvolutional capsules that can be used to utilize U-net-based architectures, namely Segcaps, to deal with segmentation tasks \cite{23}. Recently numerous works have deployed CapsNets to solve various tasks, including medical image classification \cite{24,25}, text recognition \cite{23,26,27}, and 3D point cloud classification \cite{29}.

\section{The proposed method}

{We denote an element in the $k^{th}$ capsule layer as $c_{n,d,w,h}^k \in \mathbb{R}, c^k \in \mathbb{R}^{N \times D \times W \times H}$, where $N$ is the number of capsule channels in a layer, $(W, H)$ are spatial dimensions (width and height), and $D$ is the capsule dimension. We also denote a capsule as $c_{n,:,w,h}$, when the capsule dimension is omitted. In this study, we manipulate Matlab indexing to describe a variable, where lower letters (e.g. $n, d, w, h$) are index, and capital letters (e.g. $N, D, W, H$) are the corresponding sizes.}

Before routing, a lower capsule gives a prediction for a higher capsule on the next layer by multiplication with a transformation matrix. A capsule is widely represented as a vector ($c_{n,:,w,h} \in \mathbb{R}^D$) or matrix ($c_{n,:,w,h} \in \mathbb{R}^{\sqrt{D} \times \sqrt{D}}$) \cite{1,2,3}. An advantage of the matrix form is that the number of parameters required to represent a transformation matrix is less than that required by the  vector form. For example, to perform a linear mapping from a $16-d$ vector to a $16-d$ vector, we need a  matrix of size $16 \times 16$. Alternatively, to perform a linear map between two  $4 \times 4$ matrices, we only need a $4 \times 4$ matrix. Thus, in favor of the smaller size of trainable transformation matrices, we mainly use the matrix form to represent the capsules.

\subsection{Shortcut routing}

The test of agreement \cite{1} is a selective test that determines if a higher capsule is activated by lower capsules. Suppose that the poses of capsule $A$ are represented by a matrix $T_A$. By multiplying $T_A$ by a coordinate transform $T_{AC}$, which serves as a distinct part–whole relationship between $C$ and $A$ itself, we attain a prediction for capsule $C$, $T_C=T_A T_{AC}$. Likewise, we can deduce another prediction for capsule $C$, ${T'}_{C} = T_B T_{BC}$, from capsule $B$, where $T_B$ and $T_{BC}$ denote a capsule and its part–whole relationship with capsule $C$, respectively. If capsules $A$ and $B$ comply, capsule $C$ is activated, i.e., if ${T'}_C = T_C$. The purpose of training capsule network is to discover the part–whole relationship, $T_{MN}$, between a higher capsule $N$ and a lower capsule $M$. In a computer graphics settings, $T_M$ is considered as a visual frame \cite{6} that is coordinated with an explicit viewpoint. Because the part–whole relationships are unchanged under varying viewpoints, the activated capsules remain activated from another viewpoint, while the poses $T_M$ change, according to the change of viewpoint. This property encourages capsule networks to gain an equivariant representation.

Our shortcut routing is inspired by the observation that if a higher capsule is activated, the test of agreement holds for capsules at every lower level. For example, if capsules $A$ and $B$ mutually activate capsule $C$, $T_C=T_A T_{AC}=T_B T_{BC}$, and capsule C is also involved in activating capsule $D$ with the coordinate transform $T_{CD}$, then capsules $A$ and $B$ associate in activating capsule $D$ according to the transitive property.
\begin{eqnarray}
T_D=T_C T_{CD}=T_A (T_{AC} T_{CD} )=T_B (T_{BC} T_{CD} ),
\end{eqnarray}
where $T_{BD}=T_{BC}T_{CD} (T_{AD}=T_{AC} T_{CD} )$ is the coordinate transform that relates the canonical visual entity of capsule $B (A)$ to the canonical visual entity of capsule $D$, as shown in Fig 1.
\begin{figure}[t]
	\begin{center}
		\includegraphics[width=.5\textwidth]{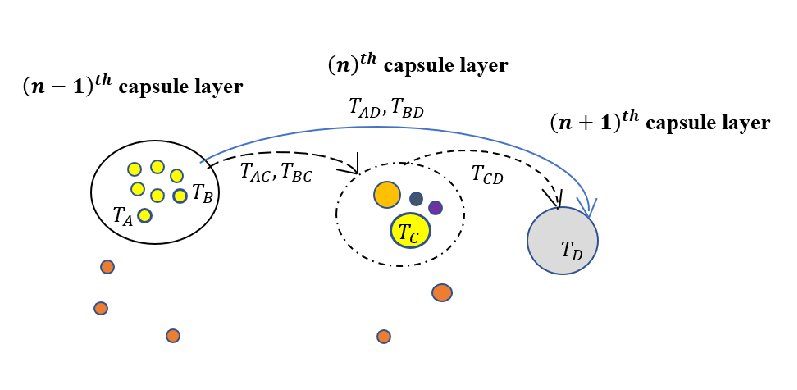}
	\end{center}
	\caption{Transitive property of the test of agreement.}
	\label{fig:1}
\end{figure}

\subsection{Building blocks}

\begin{figure}[tb]
	\begin{center}
		\includegraphics[width=.5\textwidth]{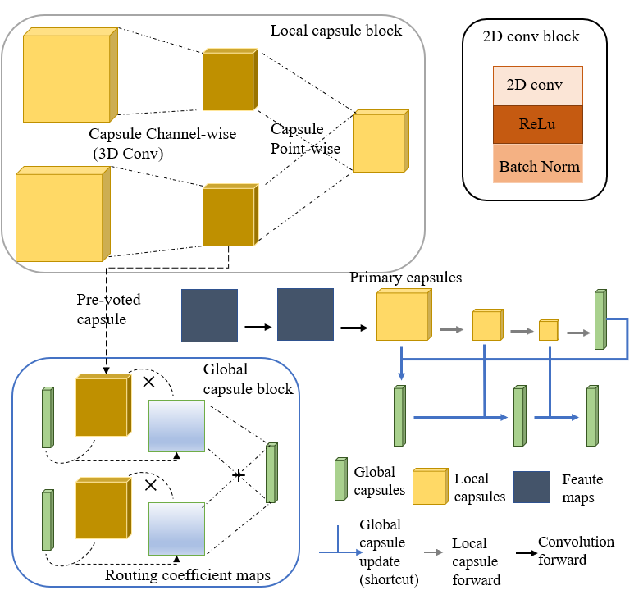}
	\end{center}
	\caption{Shortcut CapsNet: (1) backbone 2D Convolutional block: perform feature extraction. (2) local capsule block: give prediction for the capsules on the next layer. (3) Global capsule block (shortcuts:) update the global capsules by the latest global capsules and local capsules.}
	\label{fig:2}
\end{figure}

We define local capsules as capsules on intermediate layers of Capsnet, and global capsules (class capsules) as capsules on the last layer, Fig 2. The overall framework comprises two modules: local capsule block and global capsule block.

\textbf{Local capsule block} (locapblock): {The input of the locapblock is a local capsule layer, and the output is the next capsule layer, and pre-voted capsules.} This block employs depth-wise capsule convolution as its core component. Precisely, we apply a modified 3D convolution to the channel-wise approach \cite{11}. {The modified 3D convolution is a version that compatible with capsules in matrix form, where each capsule will be multiplied by a trainable transformation matrix. To achieve complete predictions for higher capsules, we apply ordinary 3D convolution on the pre-voted capsules in point-wise approach. In summary, a higher capsule is formed by a weighted combination of the lower capsules within a kernel, where the weights (routing coefficients) are shared on a channel, as explained in Eq 2.}
\begin{eqnarray}
c_{m,:,i,j}^{l+1} = \sum_n^N{r_n p_{m,:,i,j}},
\end{eqnarray}
where
\begin{eqnarray}
p_{m,:,i,j}=\sum_{w=i-k}^{i+k}\sum_{h=j-k}^{j+k}{T_{m,n, : ,w,h}^{local}\cdot c_{n,:,w,h}^l}
\end{eqnarray}
is a pre-voted capsule, $c_{n,:,w,h}^l \in \mathbb{R}^{\sqrt{D} \times \sqrt{D}}$ is a capsule at layer $l$ on the $n^{th}$ channel and at position $(w,h), T_{m,n, :, w,h}^{local} \in \mathbb{R}^{\sqrt{D} \times \sqrt{D}}$ is a transformation matrix that predicts a higher capsule, which are shared over $(i,j)$, and $"\cdot"$ denotes multiplication between matrices. Additionally, $r_n \in \mathbb{R}$ is a trainable routing coefficient on channel $n$ and differs among channels.

\textbf{Global capsule block} (glocapblock): The global block is the core operating block of subsequent steps to update the global capsules. The inputs to the glocapblock are the pre-voted capsules and the latest global capsules. We multiply each pre-voted capsule by $n$ transformation matrices to get candidates $v_{m,n,:,i,j} \in \mathbb{R}^{\sqrt{D} \times \sqrt{D}}$ of the corresponding global capsule:
\begin{eqnarray}
v_{m,n,:,i,j}= T_{n,:,i,j}^{global} \cdot p_{m,:,i,j}^l,
\end{eqnarray}
where $p_{m,:,i,j}^l$ is a pre-voted capsule that is derived in locapblock, $T_{n,:,i,j}^{global}$ is a trainable transformation matrix; which is shared among all capsules $(i,j)$ in a channel $n$. Subsequently, we update the global capsules based on the routing coefficients:
\begin{eqnarray}
g_{m,:}^t=\sum_{n}^{N}{\sum_{i}^{W}\sum_{j}^{H}{r_{m,n,i,j}v_{m,n,:,i,j}}},
\end{eqnarray}
where $g_{m,:}^t \in \mathbb{R}^{\sqrt{D} \times \sqrt{D}}$ is an updated global capsule, and $r_{m,n,i,j} \in \mathbb{R}$ is a routing coefficient, which is explained in the next section. Note that the routing algorithm is only implemented in the glocapblock.

\textbf{Unified architecture:} {Firstly, a batch of images are fed into a backbone network. The output from backbone network is then reshaped into a $5-rank$ tensor with the volume $(b, n, d, w, h)$, namely primary capsule, where $n$ is the number of capsule channels, $d$ is a square number representing the size of a capsule, $(w,h)$ are spatial dimensions, and $b$ is the batch size. Secondly, we employ the local capsule block to derive the next local capsule layers. The last layer in the first feed-forward pass is the initial prediction for the global capsules. Thirdly, following the depth of the network from the primary capsule layer, we update the global capsules by manipulating the glocapblock with the input area from a local capsule layer and the latest global capsules, as shown in the shortcut route in Fig. 2.}

\begin{figure}[tb]
	\begin{center}
		\includegraphics[width=.5\textwidth]{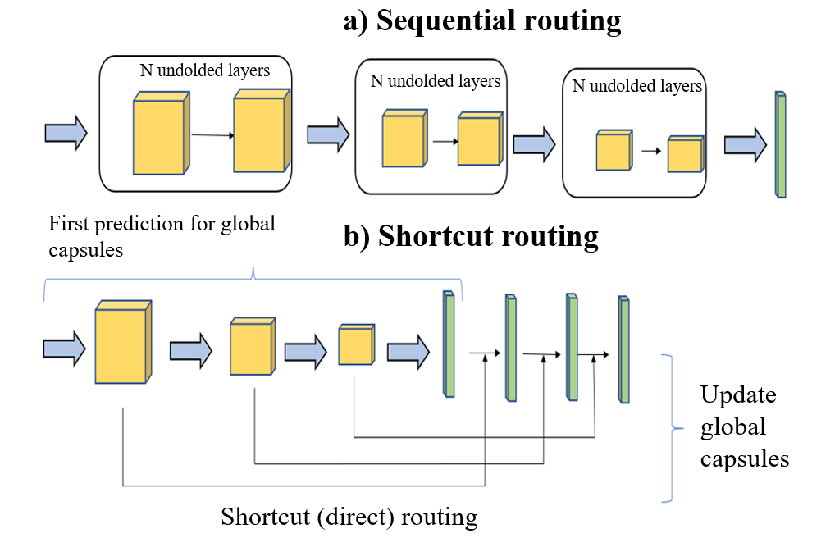}
	\end{center}
	\caption{Sequential routing \cite{2,3} and our shortcut routing. Sequential routing can unfold to the stacked capsule layers, where the number of extra layers is the number of iterations. Shortcut routing uses shortcuts to expand the depth of the network. The two architectures above have the same depth and number of iterations $(N = 2)$, but the bottom one has voting tensors of smaller size.}
	\label{fig:3}
\end{figure}

The first prediction for the global capsules in the locapblock is the initialization of a routing algorithm. The local capsules at any level directly give predictions for the global capsules via the shortcuts, which process adheres to the idea of capsules at a lower layer all being involved in activating the global capsule. {These shortcuts have the advantage of reducing the size of the voting tensors, $v \in \mathbb{R}^{M \times N \times D \times W \times H}$ in Eq. (4), where $M$ is the number of channels (or styles of capsule), $W$ and $H$ are spatial sizes, $D$ is capsule size, and $N$ is the number of capsules are needed to predict. In the classification task, the number of global capsules is much smaller than the number of local capsules at any layer $(N^{global} << N^{local})$, so as shown in Fig. 3, skipping updating local capsules, will avoid expanding the network with large size voting tensors, and make the capsule network computationally efficient.}

\subsection{Routing coefficients}

Routing or voting coefficients are scalars that weigh the degree of contribution of the lower capsules to the higher ones. CapsNets are structured from the voting process that approximates the test of agreement. In the locapblock, the routing coefficients are trainable parameters of the 3D point-wise convolution, and these parameters are shared over capsules on the same channel. Meanwhile, the routing coefficients in the glocapblock are calculated based on correlations between the local capsules in a layer and the latest global capsules. The two well-known approaches to computing the routing coefficients are attention-based \cite{3,26,30,32}, and clustering-based \cite{2,19}. The attention-based approach considers the higher capsules as queries, and the votes from the lower capsules as keys. On the other hand, the cluster-based approach defines the higher capsules as the center of the clusters, and the correlations are distances from the votes to the centers measured by a metric, such as Euclidean distance. In this study, we introduce fuzzy weights as routing coefficients, as they are more straightforward than the mixture weights in EM routing \cite{2}. We also propose a simple way to calculate routing coefficients in the attention-based approach.

A routing coefficient $r_{m,n,i,j} \in \mathbb{R}$ is determined by voting capsules $v_{m,n,:,i,j}$ and the latest global capsules $g_{m,:}$. The voting capsules are defined using learnable transformation matrices multiplied by the pre-voted capsules, as in Eq. (4). Note that the pre-voted capsules are the result of the summation of all capsules inside a kernel, as in Eq. (3). Concisely, a routing coefficient represents the contribution of all capsules in a receptive field.

Formally, for voting capsule $v_{m,n,:,i,j}$ at position $(i,j)$ on channel n, the correlation $r_{m,n,i,j}$ between $v_{m,n,:,i,j}$ and a global capsule $g_{m,:}$ is calculated as follows: 
\begin{itemize}
	\item Attention-based approach
	\begin{eqnarray}
	a_{m,n,i,j}=v_{m,n,:,i,j}\ \cdot \ g_{m,:}
	\end{eqnarray}
	\begin{eqnarray}
	r_{m,n,i,j}=\frac{exp{\left(a_{m,n,i,j}\right)}}{\sum_{k} e x p{\left(a_{k,n,i,j}\right)}},
	\end{eqnarray}
	where $a_{m,n,i,j} \in \mathbb{R}$ is an attention score, $r_{m,n,i,j}$ is the normalized attention score, and $\sum_{m} r_{m,n,i,j}=1$.
	\item Fuzzy-based approach
	\begin{eqnarray}
	d_{m,n,i,j}= \norm{v_{m,n,:,i,j} - g_{m,:}}_2
	\end{eqnarray}
	\begin{eqnarray}
	f_{m,n,i,j}=\frac{1}{\sum_{k}\left(\frac{d_{m,n,i,j}}{d_{k,n,i,j}}\right)^\frac{2}{m_f-1}}
	\end{eqnarray}
	\begin{eqnarray}
	r_{m,n,i,j}=\ \frac{\left(f_{m,n,i,j}\right)^{m_f}}{\sum_{n,i,j}\left(f_{m,n,i,j}\right)^{m_f}},
	\end{eqnarray}
	where $f_{m,n,i,j} \in \mathbb{R}$ is the fuzzy coefficient grading the degree of contribution of the local capsules (within a kernel) to the global capsule $g_{m,:}$, and $m_f$ is a fuzzy degree that is set to $2$ in this study.
\end{itemize}
\subsection{Activation probability and loss function}
The preceding section provides materials for deriving the poses (instantiations) of a capsule from their lower counterparts. Each capsule also has its own activation probability, which demonstrates how it is activated. We follow \cite{3} to calculate the probability $prob_m$ corresponding to global capsule $g_{m,:}$ in the attention-based approach:
\begin{itemize}
	\item Attention-based  approach 
	 \begin{eqnarray}
	prob_m = \norm{squash (g_{m,:})}_2 = \frac{\norm{g_{m,:}}_2}{1 + \norm{g_{m,:}}_2}
	\end{eqnarray}	
\end{itemize}
The activation probability of a global capsule is the length of its representation vector determined by squash normalization. The nonlinear squash function is applied to all capsules at each layer in the attention-based approach.
\begin{itemize}
	\item Fuzzy-based  approach 
	\begin{eqnarray}
	\sigma_m^2= \sum_n^M{\sum_j^H{\sum_i^W{r_{r,m,n,j}}{\norm{v_{m,n,:,i,j} - g_{m,:}}_2}}}
	\end{eqnarray}
	\begin{eqnarray}
		prob_m=sigmoid(\lambda(\beta_m-ln(\sigma_m)),
	\end{eqnarray}	
\end{itemize}
where $\sigma_m^2$ measures the tightness of the $m^{th}$ cluster, $\lambda$ is hyperparameter to avoid numerical errors, and $\beta_m$ is a trainable threshold. The tighter the cluster, the closer the votes are to the center global capsule. In other words, the global capsule is activated with a high probability. Our proposed formula for calculating the activation probability is not exactly the same as that introduced in \cite{2}. The probability of activation in \cite{2} is estimated from a Gaussian mixture component and the loss when activating a capsule, while our $\sigma_m^2$ is an objective function that is minimized in the fuzzy c-means algorithm.

The objective loss used to train our network is Spread Loss  \cite{2}:
\begin{eqnarray}
L =\sum_{m\neq t}{({max{\left(0,\ margin-\left(prob_m-prob_t\right)\right)}}^2},
\end{eqnarray}
where class t is the target class, and the margin ranges from $0.2$ to $0.9$ through the training time. Spread Loss aims to maximize the gap between the activation of the target class and the activation of the other classes.

\section{Experiments}
We evaluated our model on the Mnist, affNist, and smallNorb datasets. Each dataset was split into three subsets for training, validation, and testing. The splitting portions varied by dataset. After training for $200$ epochs, the model with the lowest error rate on the validation set was used to evaluate the test set. We normalized all image ranges from $0$ to $1$. We set the batch size to $128$ and trained the model with the Adam optimizer, with a learning rate of ${10}^{-3}$ and the other parameters set to default values. We decreased the learning rate by 0.8 every 20 epochs to ensure stable convergence during late training. The parameter $\lambda$ introduced in Eq. (13) was ${10}^{-1}$, the margin in the loss function (Eq. (14)) during the first ten epochs varied from $0.2$ to $0.9$  and was then unchanged. The execution environment was an Ubuntu server with a TITAN V GPU with $12$ GB RAM. We mainly implemented our method in PyTorch. The source code is available at https://github.com/Ka0Ri/shortcut-routing.git.
\subsection{Datasets}
\begin{enumerate}
	\item Mnist: Each image in the dataset is a grayscale image with a resolution of $28 \times 28$ labeled from 0 to 9. In this experiment, the training set was divided into two subsets, one for training ($50k$ images) and the other for validation ($10k$ images). When training, we randomly translated the images in a batch by a small amount as \cite{3} suggested.
	
	\item smallNorb: The database \cite{34} contains images of 50 figures belonging to 5 categories. In this study, we followed the default settings, and the test set contained the remaining 5 instances (from 0 to 5). We randomly cropped $32 \times 32$ patches of the training images, and cropped from the center of the test images. We also imposed random brightness and contrast to the training images. This augmentation was suggested in \cite{2}. As a result, the objects in the training set differ in viewpoint from the objects in the test set.
	
	\item affNist: This dataset is extended from the Mnist dataset. The dataset was built by taking images from Mnist and applying different affine transformations to them, resulting in images with a larger size of $40 \times 40$. While the default training and validation images were made by 0-padding, the test data was created by transforming the $10k$ test cases from the original Mnist dataset. Following the training scenario \cite{3}, we trained the model with translated data ($50k$ images) by randomly shifting the padded images within $5$ pixels.
\end{enumerate}

\begin{figure}[tb]
	\begin{center}
		\includegraphics[width=.5\textwidth]{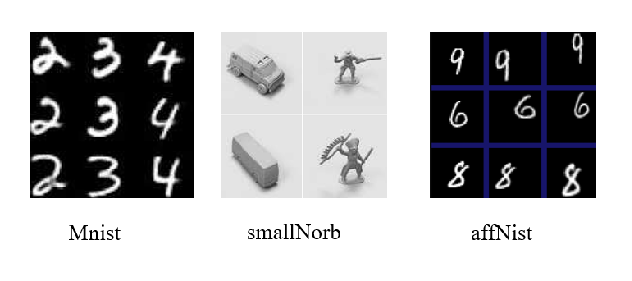}
	\end{center}
	\caption{Some images in the datasets used in the experiments}
	\label{fig:4}
\end{figure}
\subsection{The proposed architecture}
We conducted the experiments with a baseline model consisting of a backbone network of two 2D convolutional layers followed by two convcaps layers and a class capsule layer. The first convolutional layers had a kernel size of $5 \times 5$ with a stride of 2 and 64 channels. Next were $1 \times 1$ convolutional layers that formed a primary capsule layer with 8 input channels, and 128 output channels. With the above settings, the primary capsule had 8 channels, and the size of a capsule was $4\times4$. This architecture was inspired by the small model introduced in the original paper \cite{2}. After each 2D convolutional layer, we used ReLU as the activation function and then batch normalization. Subsequently, 3 consecutive locapblocks were used to calculate local capsules and derive the first predictions for global capsules. The channel-wise capsule convolution integrated inside each locapblock utilized a kernel of size $3 \times 3$ without padding. We used a stride of 2 in the first locapblock and a stride of 1 for the following locapblocks. The outputs of the channel-wise capsule convolution are pre-voted capsules. We applied point-wise 3D convolution to the pre-voted capsules with the 16 output channels. The kernel size used in the last block depends on the size of the current feature map, and reduces the size of the output feature maps to $1 \times 1$ in the fully convolutional way. 

The number of glocapblocks (shortcuts) is contingent on the depth of the model. In particular, if there are $n$ feed-forward times from the primary capsule layer to the first global capsule layer, then the number of glocapblocks is $n$. We used $n=3$ in our designated architecture, and the number of routing iterations was set to 2. We also used capsule dropout \cite{35} to make an ensemble model with a dropout probability of 0.2, and applied it to the primary capsule layer.

With the designated architecture, we aimed to construct a baseline network with the backbone is as simple as possible, so that the ability of the model would mostly depend on the capsule part. For complexity comparison, we also implemented our Capsnet with the standard configuration \cite{2,3}, where the model is wider, named as expanded model.

\subsection{Classification results}

\begin{table}[tb]
\caption{Classification result on the Mnist, affNist and smallNorb dataset (the measurement is accuracy (\%)). Details of a layer are denoted as Name: (the number of channels, kernel size, and stride size).}
\begin{tabular}{|l|l|l|l|l|l|}
\hline
\multicolumn{2}{|l|}{}                   & EM    & Fuzzy & Dynamic & Attention \\ \hline
\multicolumn{6}{|p{8cm}|}{Baseline model: Conv1: (64, 5, 2) – PrimaryCaps (8, 1, 1) – Capsconv1 (16, 3, 2) – Capsconv2 (16, 3, 2) – Classcaps (10, 3, 1)}                                          \\ \hline
\multirow{2}{*}{Mnist}     & shortcut    & 99.4  & 99.33 & 99.04   & 98.99     \\ \cline{2-6} 
                           & no shortcut & 99.37 & 99.48 & 99.43   & 99.38     \\ \hline
\multirow{2}{*}{affNist}   & shortcut    & 79.27 & 82.59 & 82.75   & 84.03     \\ \cline{2-6} 
                           & no shortcut & 66.92 & 66.85 & 65.39   & 65.75     \\ \hline
\multirow{2}{*}{smallNorb} & shortcut    & 92.16 & 91.27 & 90.26   & 89.36     \\ \cline{2-6} 
                           & no shortcut & 90.85 & 90.71 & 86.99   & 88.81     \\ \hline
\multicolumn{6}{|p{8cm}|}{Expanded model: Conv1: (64,5,2) – PrimaryCaps (32, 1, 1) – Capsconv1 (32, 3, 2) – Capsconv2 (32, 3, 2) – Classcaps (10, 3, 1)}                                          \\ \hline
\multirow{2}{*}{Mnist}     & shortcut    & 99.35 & 99.52 & 99.28   & 99.3      \\ \cline{2-6} 
                           & no shortcut & 99.20 & 99.53 & 99.44   & 99.58     \\ \hline
\multirow{2}{*}{affNist}   & shortcut    & 84.52 & 89.02 & 88.9    & 87.17     \\ \cline{2-6} 
                           & no shortcut & 73.65 & 76.95 & 70.55   & 74.06     \\ \hline
\multirow{2}{*}{smallNorb} & shortcut    & 92.69 & 93.91 & 90.31   & 91.87     \\ \cline{2-6} 
                           & no shortcut & 93.28 & 94.77 & 93.43   & 94.11     \\ \hline
\end{tabular}
\label{Table:1}
\end{table}

Table 1 lists the classification results. {With fuzzy and attention routing, we achieved competitive accuracies on all datasets. Moreover, fuzzy routing gave the highest accuracy most of the time. Additionally, our model outperforms the ordinary CapsNet on the affNist dataset.} The result of 98.17\% on the smallNorb dataset reported in \cite{2,3} is the state-of-the-art for this dataset. However, without the release of official implementation, we attained much lower accuracies using their algorithms. Eventually, the above results were lower than {those} from our proposed shortcut (92.16\% with EM routing, 91.27\% with fuzzy routing, 90.26\% with dynamic routing, and 89.36\% with attention routing).

{We addressed the trade-off between computational burden and accuracy. In particular, the number of parameters of the baseline models with shortcut routing was not large enough to obtain significant improvements overall. Although expanding learning models is a rule of thumb to deal with the underfitting problem, when working with capsule networks it seems to be impractical, because expanding a capsule network exponentially increases the number of calculations with respect to the number of parameters. Herein, our proposed shortcut routing takes advantage of expanding the model with the simplified architecture, and which when fuzzy or attention routing is integrated, is even more compact.  We have shown that expanded models with shortcut routing achieved higher accuracy compared to all baseline models without shortcut routing, even though the numbers of parameters of those models were approximate.

Although the results of the shortcut approach with the expanded model were a little inferior to those of the conventional approach, the shortcut routing was much faster than the original architecture. As suggested by the results, we expect that the proposed architecture would help build deeper capsule networks to enhance the learning ability of models.}

\subsection{Performance Analysis}

Our proposed model is computationally efficient compared to the original version \cite{2,3}. The efficiency is demonstrated in two quantities: the number of calculations and the number of trainable parameters. {When comparing the number of parameters used in each model, the baseline model with shortcut routing consists of 23k parameters}. The architecture proposed in \cite{2} has the same depth in the capsule part as our proposed model (3 capsule convolutional layers), but the number of parameters is much higher (88k) than in our model, and considerably fewer than that used by the model in \cite{3}, with 8.2M parameters . Notably, the expanded architecture using shortcut routing has 68k parameters, which is even fewer than the number of parameters in the baseline architecture without shortcuts (88k). It is substantially smaller than the original model without shortcuts (377k). Our small-scale model is the result of exploiting the depth-wise capsule convolution, shortcut routing, and transformation matrices shared in a channel.

Another quantity involved in the computational burden in CapsNets is the size of voting tensors. The votes are the predictions of the lower capsules for the higher capsules abiding by the $M-N$ correspondence, where $M (N)$ is the number of lower (higher) capsules. Routing methods consider the voting matrix as explicit input to derive the higher-level capsules and their corresponding activation probability. Theoretically, it is sufficient to compare two routing procedures based solely on the calculations in an iteration. 

A prediction from the capsule at position $(i,j)$ on channel $n$ at an intermediate layer for the global capsule $m$ is $v_{m,n,:,i,j}$. Thus, the size of the voting tensor in our implementation is $C_{out}\times C_{in}\times H\times W\times D$, where $C_{out}$ is the number of global capsules, $C_{in}$ is the number of channels of the input channel , and $W \times H$ is the spatial size of the pre-voted capsule layer. On the other hand, routing between two consecutive layers in the patch-wise approach presented in [2] yields a voting tensor of size $K\times K\times{C'}_{out}\times{C'}_{in}\times W' \times H' \times D$, where ${C'}_{out} ({C'}_{in})$ is the number of  output (input) channels, $W' \times H'$ is the spatial size of the target layer, and $K\times K$ is the size of a patch. The ratio of the size of the voting tensors in our shortcut approach and without shortcut approach is 
\begin{eqnarray}
r_{voting} = \frac{C_{out}\times C_{in}\times H\times W\times D}{K\times K\times{C'}_{out}\times{C'}_{in}\times W'\times H'\times D}, \nonumber\\
\end{eqnarray}
{where $C_{in}=C'_{in}$ is the number of channels in the current layer, and in fact $W'=W$ and $H'=H$ since the spatial size of the pre-voted capsule layer is the spatial size of the target layer in our approach, Eq. (15) can be reduced to}
\begin{eqnarray}
r_{voting} = \frac{C_{out}}{K\times K\times{C'}_{out}}.
\end{eqnarray}
We further address the computational problem in the routing algorithm by introducing two simplified ways of calculating the routing coefficients: the attention-based approach based on cosine similarity and the fuzzy-based approach based on fuzzy coefficients. To compare these approaches, we calculated the number of FLOPS theoretically required in a routing iteration in the fuzzy-based, attention-based, and EM-based algorithms \cite{2}. {Denoting $Q$ as the total size of the voting tensor, in approximation, the attention-based approach requires $4Q$ FLOPS, and the fuzzy-based approach requires $7Q$ FLOPS, whereas the EM-based approach requires $10Q$ FLOPS}. These results show that our fuzzy-based, and attention-based algorithms are 1.42 times, and 2.5 times more efficient than the EM-based algorithm, respectively. Moreover, the size Q of the voting tensor in our proposed algorithms is much smaller than that in EM-based routing (Eq. (16)), thereby making our model overall computationally efficient.

In practice, the training time of fuzzy-based routing and attention-based routing is shorter than that of EM-based routing, given images of size $28 \times 28$; {Table 2 provides the details of consumptions}. Moreover, the memory consumed by fuzzy-based routing and attention-based routing is also lower than that consumed by EM-based routing. These results are reported in the training phase with the same configuration. Table 2 summarizes the above performance analysis.

\begin{table}[tb]
\caption{Execution time (ms), memory usage (MB), and the number of training parameters for each method. }
\begin{tabular}{|l|l|l|l|l|l|}
\hline
\multicolumn{2}{|l|}{}                  & EM    & Fuzzy & Dynamic & Attention \\ \hline
\multicolumn{6}{|l|}{Baseline model}                                          \\ \hline
\multirow{3}{*}{shortcut} & No. of parameters   & \multicolumn{4}{l|}{23k}            \\ \cline{2-6} 
                  & Execution time & 1.236 & 1.001 & 0.684   & 0.672     \\ \cline{2-6} 
                  & Memory usage  & 1171  & 1169  & 1167    & 1167      \\ \hline
\multirow{3}{*}{no shortcut} & No. of parameters   & \multicolumn{4}{l|}{88k}            \\ \cline{2-6} 
                  & Execution time      & 1.501 & 1.296 & 1.224   & 0.696     \\ \cline{2-6} 
                  & Memory usage        & 1249  & 1229  & 1225    & 1205      \\ \hline
\multicolumn{6}{|l|}{Expanded model}                                          \\ \hline
\multirow{3}{*}{shortcut} & No. of parameters   & \multicolumn{4}{l|}{68k}            \\ \cline{2-6} 
                  & Execution time      & 1.386 & 1.252 & 1.236   & 1.235     \\ \cline{2-6} 
                  & Memory usage        & 1193  & 1185  & 1175    & 1175      \\ \hline
\multirow{3}{*}{no shortcut} & No. of parameters   & \multicolumn{4}{l|}{377k}           \\ \cline{2-6} 
                  & Execution time      & 6.636 & 5.1   & 4.248   & 4.008     \\ \cline{2-6} 
                  & Memory usage        & 1589  & 1465  & 1441    & 1375      \\ \hline
\end{tabular}
\label{Table:2}
\end{table}

\section{Conclusions}
Our study introduces a new routing path for capsule networks, namely shortcut routing. To this end, we developed two modules that were integrated into the architecture of CapsNets. First, the local routing block from the depth-wise convolution with a modification is equivalent to matrix multiplication. Second, the global capsule block helps in explicit routing from local capsules to global capsules with the shortcuts. Additionally, we presented fuzzy coefficients as a computationally efficient voting method for deducing higher-level capsules from the lower ones. The proposed framework achieved on-par performance with the original benchmark, while reducing in the complexity of model architecture, as well as of routing algorithm. This study therefore contributes a united framework that shows promise for further research into capsule networks.


\begin{thebibliography}{99}
\bibitem{1}
G. E. Hinton, A. Krizhvesky and S. D. Wang, Transforming Auto-Encoders, International Conference on Artificial Neural Networks - ICANN 2011, Berlin, 2011.

\bibitem{2}
G. E. Hinton, S. Sabour and N. Frosst, Matrix Capsules With EM Routing, International Conference on Learning Representations, 2018.

\bibitem{3}
S. Sabour, N. Frosst and G. E. Hinton, Dynamic Routing Between Capsules, Advances in neural information processing systems, pp. 3856-3866, 2017.

\bibitem{4}
A. Kosiorek, S. Sabour, Y. W. Teh and G. E. Hinton, Stacked capsule autoencoders, Advances in Neural Information Processing Systems, pp. 15486-15496, 2019.

\bibitem{6}
G. E. Hinton, Z. Ghahramani and W. Y. Teh, Learning to Parse Images, Advances in Neural Information Processing Systems 12, 1999.

\bibitem{7}
R. S. Zemel, C. M. Mozer and G. E. Hinton, TRAFFIC: Recognizing Objects Using Hierarchical Reference Frame Transformations, Advances in Neural Information Processing Systems 2, 1989.

\bibitem{5}
I. Goodfellow, Y. Bengio and A. Courville, Manifold Learning, in Deep Learning,  The MIT Press, pp. 156-159, London, 2017.

\bibitem{8} 
Y. Bengio, A. Courville and P. Vincent, Representation learning: A review and new perspectives, IEEE transactions on pattern analysis and machine intelligence, vol. 35, no. 8, pp. 1798-1828, 2013.

\bibitem{10} 
K. He, X. Zhang, S. Ren and J. Sun, Deep residual learning for image recognition, Proceedings of the IEEE conference on computer vision and pattern recognition, 2016.



\bibitem{12} 
T. D. Kulkarni, W. Whitney, P. Kohli and J. B. Tenenbaum, Deep Convolutional Inverse Graphics Network, Advances in neural information processing systems, pp. 2539-2547, 2015.

\bibitem{13} 
M. Jaderberg, K. Simonyan, and A. Zisserman, Spatial transformer networks, Advances in neural information processing systems, pp. 2017-2025, 2015.

\bibitem{14} 
D. E. Worrall, S. J. Garbin, D. Turmukhambetov and G. Brostow, Harmonic Networks: Deep Translation and Rotation Equivariance, in Proceedings of the IEEE Conference on Computer Vision and Pattern Recognition, 2017.

\bibitem{15}
I. Goodfellow, Y. Bengio and A. Courville, Adversarial Training, in Deep Learning, London, The MIT Press, pp. 261-263, 2017.

\bibitem{16} 
I. J. Goodfellow, J. Shlens and C. Szegedy, Explaining and Harnessing Adversarial Examples, arXiv preprint arXiv:1412.6572, 2014.

\bibitem{17} 
J. E. Lenssen, M. Fey and P. Libuschewski, Group equivariant capsule networks, Advances in Neural Information Processing Systems, pp. 8844-8853, 2018.

\bibitem{18} 
T. S. Cohen and M. Welling, Group equivariant convolutional networks, International conference on machine learning, 2016.

\bibitem{30} 
J. Choi, H. Seo, S. Im and M. Kang, Attention Routing Between Capsules, Proceedings of the IEEE International Conference on Computer Vision Workshops, 2019.

\bibitem{31}  
H. Li, X. Guo, B. Dai, W. Ouyang and X. Wang, Neural Network Encapsulation, Proceedings of the European Conference on Computer Vision (ECCV), 2018.

\bibitem{32}  
K. Ahmed and L. Torresani, STAR-CAPS: Capsule Networks with Straight-Through Attentive Routing, Advances in Neural Information Processing Systems, pp. 9098-9107, 2019.

\bibitem{19} 
S. Zhang, W. Zhao, X. Wu and Q. Zhou, Fast dynamic routing based on weighted kernel density estimation, International Symposium on Artificial Intelligence and Robotics, pp. 301-309, 2018.

\bibitem{20} 
K. Duarte, Y. S. Rawat and M. Shah, VideoCapsuleNet: A Simplified Network for Action Detection, arXiv preprint arXiv:1805.08162, 2018. 
\bibitem{21} 
J. Rajasegaran, V. Jayasundara, S. Jayasekara, H. Jayasekara, S. Seneviratne and R. Rodrigo, DeepCaps: Going Deeper with Capsule Networks, Proceedings of the IEEE Conference on Computer Vision and Pattern Recognition, 2019.
 
\bibitem{22} 
L. Zhang, M. Edraki and G. J. Qi, CapProNet: Deep feature learning via orthogonal projections onto capsule subspaces, Advances in Neural Information Processing Systems, pp. 5814-5823, 2018.

\bibitem{23} 
R. LaLonde and U. Bagci, Capsules for object segmentation, arXiv preprint arXiv:1804.04241, 2018.

\bibitem{24} 
P. Afshar, A. Mohammadi and P. Konstantinos, Brain tumor type classification via capsule networks, in 2018 25th IEEE International Conference on Image Processing (ICIP), 2018.
 
\bibitem{25} 
T. Iesmantas and R. Alzbutas, Convolutional capsule network for classification of breast cancer histology images, International Conference Image Analysis and Recognition, 2018.

\bibitem{26} 
X. Zhang, P. Li, W. Jia and H. Zhao, Multi-labeled Relation Extraction with Attentive Capsule Network, arXiv preprint arXiv:1811.04354, 2018.
 
\bibitem{27} 
W. Zhao, J. Ye, M. Yang, Z. Lei, S. Zhang and Z. Zhao, Investigating capsule networks with dynamic routing for text classification, arXiv preprint arXiv:1804.00538, 2018.

\bibitem{29} 
Y. Zhao, T. Birdal, H. Deng and F. Tombari, 3D Point Capsule Networks, Proceedings of the IEEE Conference on Computer Vision and Pattern Recognition, 2019.

\bibitem{11} 
A. G. Howard, M. Zhu, B. Chen, D. Kalenichenko, W. Wang, T. Weyand, M. Andreetto and H. Adam, MobileNets: Efficient Convolutional Neural Networks for Mobile Vision Applications, arXiv preprint arXiv:1704.04861, 2017.

\bibitem{34} 
L. Yann, J. F. Huan and L. Bottou, Learning Methods for Generic Object Recognition with Invariance to Pose and Lighting, IEEE Computer Society Conference on Computer Vision and Pattern Recognition, 2004. 
\bibitem{35} 
C. Xiang, L. Zhang, Y. Tang, W. Zou and C. Xu, "MS-CapsNet: A novel multi-scale capsule network," IEEE Signal Processing Letters, vol. 25, no. 12, pp. 1850-1854, 2018. 


\end{thebibliography}

\profile[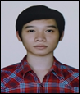]{Dang Thanh Vu}{received B.S. degrees in Mathematics and Computer Science from Vietnam National University in 2018. He is now a M.S. student in the Department of ICT Convergence System Engineering, Chonnam National University, Rep. of Korea. His research interests include statistical methods, and machine learning.}
\profile[author2]{Vo Hoang Trong}{received M.S. degree in Electronics Engineering from Chonnam National University in 2020. He is now a Ph.D. student in the Department of ICT Convergence System Engineering, Chonnam National University, Rep. of Korea. His research interests include image processing, and deep learning.}
\profile[author3]{Yu Gwanghyun}{received M.S. degree in Electronics Engineering from Chonnam National University in 2018. He is now a Ph.D. student in the Department of ICT Convergence System Engineering, Chonnam National University, Rep. of Korea. His research interests include signal processing, and deep learning.}
\profile[author4]{Kim Jin Young}{received his Ph.D. degree in Electronics Engineering from Seoul National University, Rep. of Korea. From 1993 to 1994, he worked on speech synthesis at Korea Telecom. Since 1995, he has been a professor in the Department of ICT Convergence System Engineering, Chonnam National University, Rep. of Korea. His research interests include speech signal processing, audio-visual
speech processing, and deep learning-based signal processing}
\end{document}